\documentclass[journal]{IEEEtran}
\ifCLASSOPTIONcompsoc
  \usepackage[nocompress]{cite}
\else
  \usepackage{cite}
\fi
\ifCLASSINFOpdf
\else
\fi
\usepackage[utf8]{inputenc} 
\usepackage[T1]{fontenc}    
\usepackage{url}            
\usepackage{amsfonts}       
\usepackage{nicefrac}       
\usepackage{microtype}      
\usepackage{graphicx}
\usepackage{mathtools}
\usepackage{multirow}
\usepackage{subfigure}
\usepackage{tabularx}
\usepackage{amsmath,amsfonts,amssymb}
\usepackage{wrapfig,lipsum,booktabs}
\usepackage{xcolor}
\usepackage[pagebackref=true,breaklinks=true,letterpaper=true,colorlinks,bookmarks=false,citecolor=cyan]{hyperref}
\usepackage{balance}
\usepackage{bbm}
\def\etal{\textit{et al.}}

\def\vs{\textit{vs.}}

\usepackage{tikz}

\usepackage{enumitem}

\begin{document}
\setcounter{page}{1}
\title{Geo-ConvGRU: Geographically Masked Convolutional Gated Recurrent Unit for Bird’s-Eye View Segmentation
}
%
%
%
%

\author{Guanglei Yang$^\dagger$,
Yongqiang Zhang$^\dagger$,
Wanlong Li$^*$,
Yu Tang,
Weize Shang,
Feng Wen,
Hongbo Zhang,
Mingli Ding
\IEEEcompsocitemizethanks{
\IEEEcompsocthanksitem Guanglei Yang,Yongqiang Zhang and Mingli Ding are with the School of Instrument Science
and Engineering, Harbin Institute of Technology (HIT), Harbin, China. E-mail: \{yangguanglei, zhangyongqiang, dingml\}@hit.edu.cn\protect
\IEEEcompsocthanksitem Wanlong Li, Wen Feng and Hongbo Zhang are with Huawei Noah's Ark Lab, Huawei inc., Beijing, China. E-mail: \{liwanlong,tangyu17,feng.wen,zhanghongbo888\}
  @huawei.com \protect
\IEEEcompsocthanksitem Weize Shang is the school of Astronautics, Beihang University, Beijing, China. E-mail:  \{shangweize\}@buaa.edu.cn \protect  
  }

\thanks{$^\dagger$ Equal contribution.\protect}
\thanks{$^*$ Corresponding author.\protect}

}


%
%

\markboth{Submitted to IEEE Transactions on Intelligent Vehicles}%
{Shell \MakeLowercase{\textit{et al.}}: Bare Demo of IEEEtran.cls for Computer Society Journals}
%



\IEEEtitleabstractindextext{%
\begin{abstract}

Convolutional Neural Networks (CNNs) have significantly impacted various computer vision tasks, however, they inherently struggle to model long-range dependencies explicitly due to the localized nature of convolution operations. Although Transformers have addressed limitations in long-range dependencies for the spatial dimension, the temporal dimension remains underexplored. In this paper, we first highlight that 3D CNNs exhibit limitations in capturing long-range temporal dependencies. Though Transformers mitigate spatial dimension issues, they result in a considerable increase in parameter and processing speed reduction. To overcome these challenges, we introduce a simple yet effective module, Geographically Masked Convolutional Gated Recurrent Unit (Geo-ConvGRU), tailored for Bird's-Eye View segmentation. Specifically, we substitute the 3D CNN layers with ConvGRU in the temporal module to bolster the capacity of networks for handling temporal dependencies. Additionally, we integrate a geographical mask into the Convolutional Gated Recurrent Unit to suppress noise introduced by the temporal module. Comprehensive experiments conducted on the NuScenes dataset substantiate the merits of the proposed Geo-ConvGRU, revealing that our approach attains state-of-the-art performance in Bird's-Eye View segmentation.

\end{abstract}
}

\maketitle

\IEEEdisplaynontitleabstractindextext

%
\IEEEpeerreviewmaketitle

\section{Introduction}


Over the past decade, convolutional neural networks have emerged as the primary methodology for addressing fundamental and challenging computer vision tasks. Since the seminal work of~\cite{dosovitskiy2020image} in 2D computer vision, lack of spatial dimension long-range dependencies in CNNs has garnered increasing attention. However, for autonomous driving tasks such as Bird's-Eye View (BEV) semantic segmentation and future prediction, the limited temporal receptive field of 3D CNNs has often been overlooked by prior methods~\cite{zhu2017overview,philion2020lift, hu2021fiery, hu2022stp3,cao2022future,chen2022milestones,zhang2023controlvideo}.

\begin{figure}[!t]
    \centering
    \includegraphics[width=0.95\linewidth]{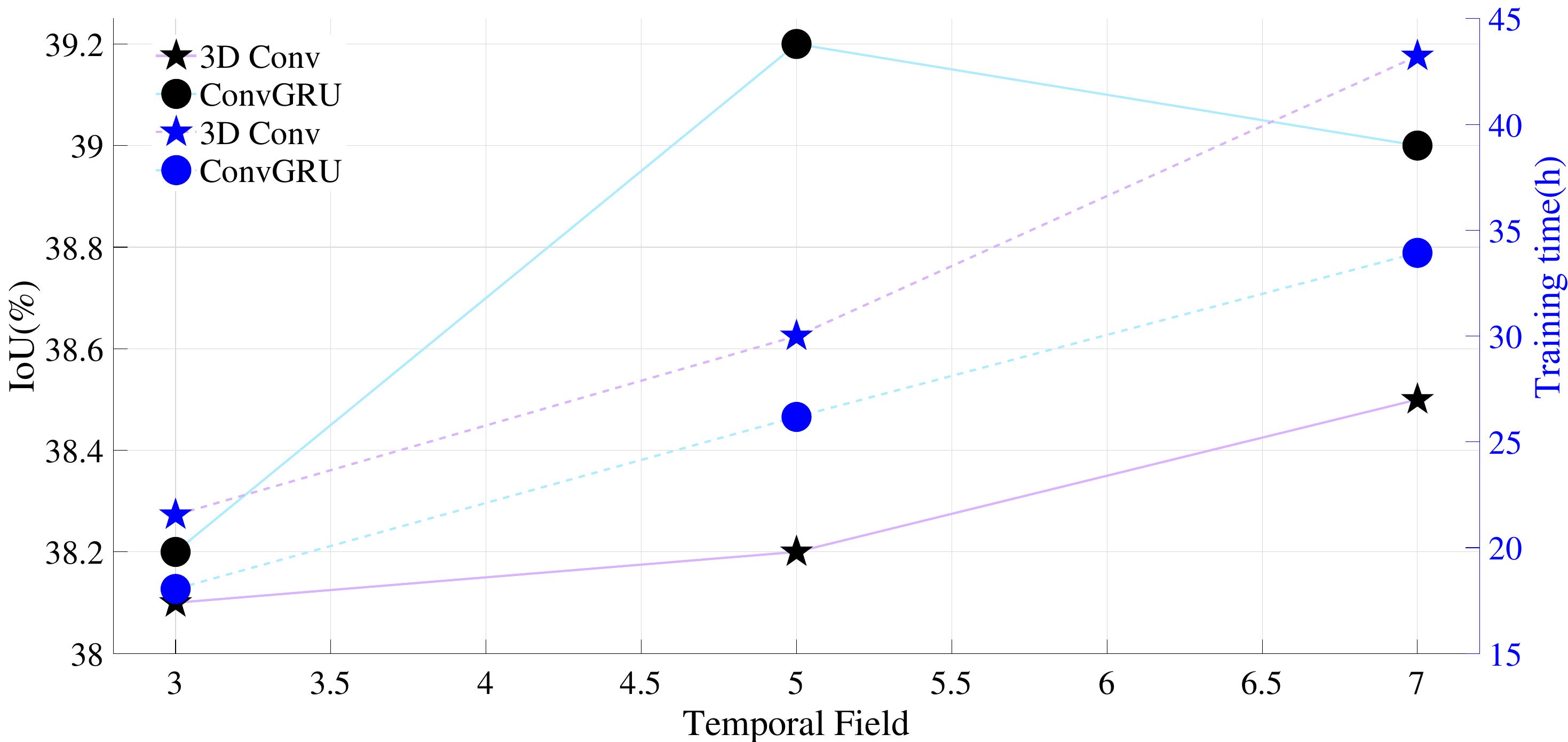}
    \caption{{
    Performance (IoU) \vs~efficiency (Training Time) with the different temporal filed on bird's-eye view semantic segmentation.}}
    \label{fig:temporal}
\vspace{-0.4cm}
\end{figure}

BEV segmentation, the task of assigning a semantic label to each pixel in BEV space, is crucial for autonomous driving, enabling safe and accurate navigation of a vehicle through 3D space. As demonstrated by~\cite{hu2021fiery}, incorporating a temporal module to extract Spatial-temporal representations from several frames significantly improves model performance. According to our experiment (shown in Fig.\ref{fig:temporal}), when the temporal field increases (adding more frames to the temporal module), the performance of the 3D CNN model does not achieve a reasonable boost. To address this limitation,\cite{liu2022petrv2} and \cite{li2022bevformer} employ Spatial-temporal Transformers for large temporal respective field. Autonomous driving tasks require efficient and reliable strategies for incorporating Transformer modules. However, current approaches have some drawbacks that hinder their effectiveness. For example, the serial fusion between current and previous frames results in low computational efficiency for multiple frames. Additionally, the inclusion of a Transformer module significantly increases the model's parameters and GPU memory consumption, making it is challenging to meet real-time requirements for autonomous driving tasks. As such, there is a need to explore alternative approaches that can address these issues and ensure optimal performance in autonomous driving scenarios.

To capture long-range temporal dependencies efficiently, in this paper, we employ a convolutional gated recurrent unit (ConvGRU) as the temporal module~\cite{ballas2016delving} for BEV segmentation. This module is used to explore temporal variations from visual percepts at different spatial resolutions.
Our network leverages ConvGRU to extract robust temporal features and share parameters across input spatial locations. This approach replaces a fully connected layer with a convolution, significantly reducing memory requirements.
However, some noises are introduced during the fusion of spatial and temporal features in ConvGRU, making it is challenging for the model to predict the accurate spatial location of moving neighborhood cars. To mitigate this issue, we further generate a geographical mask using geographical containment.
Our experimental results show that adding a geographical mask further enhances the model's performance. Overall, we are continuously working to improve our network's performance and enhance the accuracy of our predictions.


Compared to previous state-of-the-art methods, our Geo-ConvGRU achieves \textbf{1.3\%}, \textbf{0.9\%}, and \textbf{0.8\%} improvements in BEV semantic segmentation, future instance segmentation, and perceived map prediction, respectively. These improvements validate the effectiveness of our method for autonomous driving tasks.

Our contributions can be summarized as follows:

\begin{itemize}
\vspace{.05in}
\item We demonstrate that 3D convolutional neural networks exhibit limitations in long-range dependencies along the temporal dimension, as evidenced by our experiments. To address this issue, we adopt the convolutional gated recurrent unit (ConvGRU) as the temporal module.
\vspace{.05in}
\item We propose a novel, simple, but effective geographical mask using geographical containment, which can alleviate the overfitting problem at moving pixels caused by the fusion of spatial and temporal features.
\vspace{.05in}
\item Extensive experiments conducted on the NuScenes dataset substantiate the effectiveness of our Geo-ConvGRU framework, which outperforms state-of-the-art methods.

\end{itemize}

\begin{figure*}[t]
    \centering
    \includegraphics[width=0.95\linewidth]{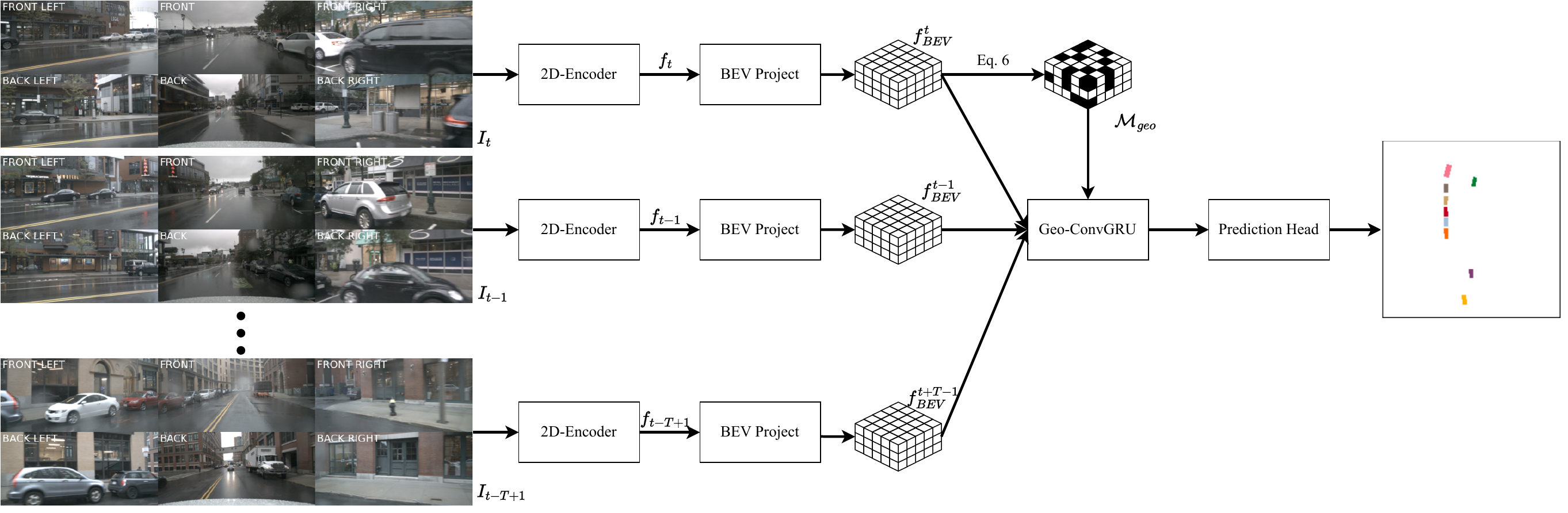}
    \caption{The overview of a segmentation model for BEV segmentation. 
    }
    \label{fig:idea}
\end{figure*}

\section{Related Work}

With the development of 3D recognition datasets~\cite{caesar2020nuscenes, cordts2016cityscapes, geiger2012we}, numerous works have focused on perception in the BEV view. Typically, BEV segmentation uses multi-view camera images as inputs and assigns a semantic label to each pixel in the BEV space. VPN~\cite{pan2020cross} first introduces cross-view semantic segmentation to understand the surrounding environment and proposes a view parsing network to fuse multi-view features. LSS~\cite{philion2020lift} lifts each camera image individually into a frustum of features for each camera and splats all frustums into a bird's-eye-view grid. Following LSS, Fiery~\cite{hu2021fiery} also employs depth prediction to project camera images into BEV space. The distinction in Fiery is adding a temporal module to extract more temporal information than LSS. According to Fiery, the temporal module has been proven as an effective way to enhance the model's performance in BEV segmentation. 

Similarly, BEVFormer~\cite{li2022bevformer} aims to exploit both spatial and temporal information via spatio-temporal transformers. Moreover, Zhou \etal ~\cite{zhou2022cross} propose adding cross-view transformer layers to infer map-view for semantic segmentation. PERT~\cite{liu2022petr, liu2022petrv2} extends the 3D position embedding in the original transformer for temporal modeling. As expected, incorporating a Transformer significantly boosts the model's accuracy, but at the cost of increased model parameter size and GPU memory consumption. These drawbacks conflict with the real-time requirements for autonomous driving tasks. To address this issue, ST-P3~\cite{hu2022stp3} proposes a temporal-based refinement unit in the planning decoder. However, these previous methods fail to consider the limitations of 3D CNNs in the temporal field. 

In this paper, we adopt a convolutional gated recurrent unit (ConvGRU) as the temporal module to extend the model's temporal receptive field for the task of Bird’s-Eye View segmentation. ConvGRU~\cite{ballas2016delving} is an RNN model designed to process a sequence of inputs. It has been widely applied in various tasks, such as machine translation and image/video caption generation. Moreover, ConvGRU has empirically demonstrated its ability to model long-term temporal dependency~\cite{tran2018closer, feichtenhofer2017spatiotemporal, ji2021full}. Compared to Long Short Term Memory (LSTM)~\cite{hochreiter1997long}, the main advantage of ConvGRU is its lower memory requirement~\cite{chung2014empirical}. As shown in Fig~\ref{fig:temporal}, ConvGRU can significantly improve performance with more frames added while not leading to a substantial increase in time consumption.

\section{Methodology}

In this section, we introduce our Geo-ConvGRU for BEV segmentation from multiple camera views. At temporal field $T$, we take camera inputs $o_t,...,o_{t-T+1}$ to BEV space using a self-supervised depth probability distribution over pixels and camera's parameters. For each frame camera input $o_i$, it comprises a set of $N$ monocular views $(I_n,K_n,R_n,t_n)_{n=1}^N$, including the input image $I_n\in\mathbb{R}^{H\times H \times 3}$, camera intrinsics $K_n\in \mathbb{R}^{3\times 3}$, and extrinsic rotation $R_n\in\mathbb{R}^{3\times3}$ and translation $t_n\in\mathbb{R}^3$ relative to the ego-vehicle's center. We aim to train an efficient network for BEV segmentation, extracting features from both spatial and temporal dimensions to predict a semantic/instance segmentation mask $\hat{y}\in {0,1}^{H\times W \times C}$ in the BEV space. In all experiments, the number of camera views $N$ is set to 6.

We first introduce the framework overview of Geo-ConvGRU in Section~\ref{sec:overview}. Then, in Section~\ref{sec:gru}, we describe the details of the geographically masked convolutional gated recurrent unit.

\subsection{Framework Overview}
\label{sec:overview}

Figure~\ref{fig:idea} shows the overview of the proposed method. The entire segmentation model comprises a backbone, BEV projection module, Geo-ConvGRU module, and prediction head. The backbone is a fine-tuning network that extracts the initial 2D features  from a sequence of camera inputs. Similar to~\cite{hu2021fiery} and \cite{philion2020lift}, we use the same BEV projection process to transform 2D features into BEV space. The difference between our approach and~\cite{hu2021fiery,philion2020lift} in the BEV projection process is the additional output, the geographical mask $\mathcal{M}_{geo}$, derived from camera's intrinsic and extrinsic parameters. Section~\ref{sec:gru} provides more details on generating the geographical mask $\mathcal{M}_{geo}$. 

The BEV features are fed to our Geo-ConvGRU module to extract temporal representation. Simultaneously, the output of ConvGRU undergoes element-wise multiplication with the geographical mask $\mathcal{M}_{geo}$ {to suppress the influence of non-visible pixels.} According to our experimental results, geographical mask weighting with features can help remove spatial noise due to the addition of temporal modules. Finally, a prediction head is used to output the final BEV segmentation results.

\subsection{Convolutional Gated Recurrent Unit}
\label{sec:gru}

\begin{figure}[t]
    \centering
    \includegraphics[width=0.95\linewidth]{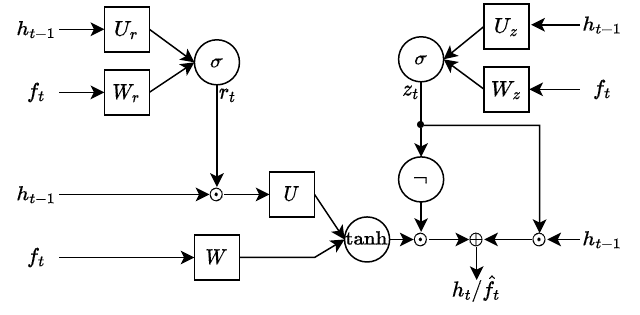}
    \caption{A example for ConvGRU unit.The $\neg$ denotes NOT process.}
    \label{fig:gru}
\end{figure}

In this paper, we adapt the
convolutional gated recurrent unit (ConvGRU) to enhance the model's long-term temporal dependency for BEV segmentation. Contrasting with huge parameters of the other previous spatial-temporal module, ConvGRU can reach the reasonable trade-off between parameters and performance.

Fig.~\ref{fig:gru} depicts an example of ConvGRU unit.
The output is computed as follows:
\begin{equation}
    z_t=\sigma(W_z\ast f_t +U_z\ast h_{t-1}),
\end{equation}
\begin{equation}
    r_t=\sigma(W_r\ast f_t +U_r\ast h_{t-1}),
\end{equation}
\begin{equation}
    \widetilde{h}_t=\tanh(W\ast f_t +U\ast(r_t\odot h_{t-1})),
\end{equation}
\begin{equation}
    h_t/\hat{f}_t=(1- z_t)h_{t-1}+z_t\widetilde{h}_t,
\end{equation}
where $\ast$ denotes a convolution operation. In these formulations, $W, W_z, W_r$, and $U, U_z, U_r$ are 2D-convolutional kernels. Moreover, $r_t$, $z_t$ are reset gate and updated gate, respectively.

\subsection{Geographically Masked Convolutional Gated Recurrent Unit}

{A comprehensive ablation study reveals a subtle inclination of the model to occasionally generate incorrect predictions concerning moving cars.} This is due to the activation of certain invalid voxels through the temporal module, which results in no 2D pixel projections during the BEV projection. {Nonetheless, the development of a distinctive geographical mask has been successful. This mask enhances the impact of valid voxels while concurrently suppressing others, thus systematically addressing the mentioned issue.}

\begin{figure}[h]
    \centering
    \includegraphics[width=0.95\linewidth]{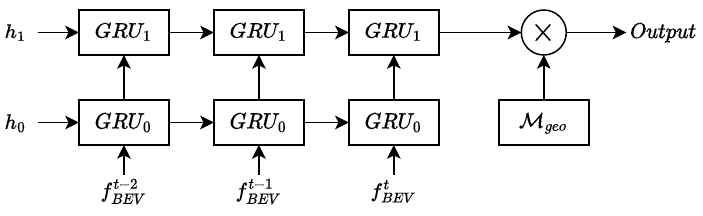}
    \caption{A example for our Geo-ConvGRU module. $\mathcal{M}_{geo}$ denotes the geographical mask and $\otimes$ means the element-wise product operation.}
    \label{fig:geoconvgru}
\end{figure}

The overview of proposed Geo-ConvGRU is shown in Fig~\ref{fig:geoconvgru}.
The entire Geo-ConvGRU consists of two parts: a ConvGRU module and a geographical mask weighting operation. In detail, the number of ConvGRU units and the temporal field (T) are set to 2 and 3, respectively.

Given the camera dimension $\mathbb{R}^{[h,w,d]}$ and BEV dimension $\mathbb{R}^{[H,W,D]}$, the operation projecting from 2D space to BEV space can be defined as:
\begin{equation}
    \hat{P}_{BEV}=RkP_{2d}+t,
\end{equation}
where $K, R, t$ represent the camera intrinsics, extrinsic rotation, and extrinsic translation.

{For the geometry mask $\mathcal{M}_{geo}$, 
the value of the voxel $v$ is set to 1 if the point in the BEV space is visible in any camera.} We denote these voxels as valid, while others are considered invalid in the BEV space. This process can be described as:
\begin{equation}
\mathcal{M}_{geo}=\left\{\begin{matrix}
1 & \exists p \in v\ \& \ p\in \hat{P}_{BEV}\\
\varepsilon & otherwise \\
\end{matrix}\right.
\end{equation}
where $\varepsilon$ is a small value for numerical stability. In our experiment, $\varepsilon$ is set to 0.1 for all tasks.

Finally, the whole Geo-ConvGRU module can be summarised as:
\begin{equation}
    \hat{f}_{BEV}=\mathcal{M}_{geo}\psi(f_{BEV}),
\end{equation}
where $\psi$ represents the ConvGRU function.

\begin{figure*}[!t]
\centering
\subfigure[Image]{
\begin{minipage}{0.39\linewidth}
        \centering
        \includegraphics[width=0.993\textwidth,height=0.9in]{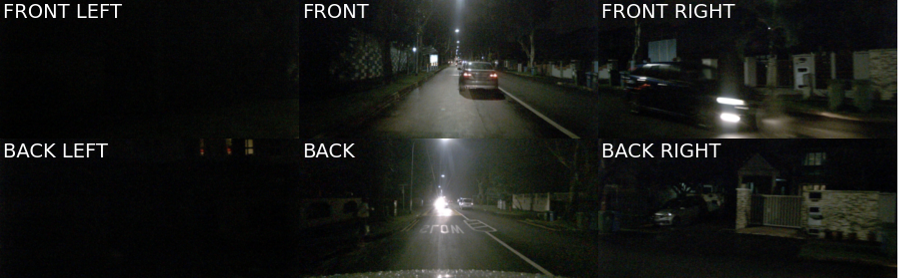}\\
        \includegraphics[width=0.993\textwidth,height=0.9in]{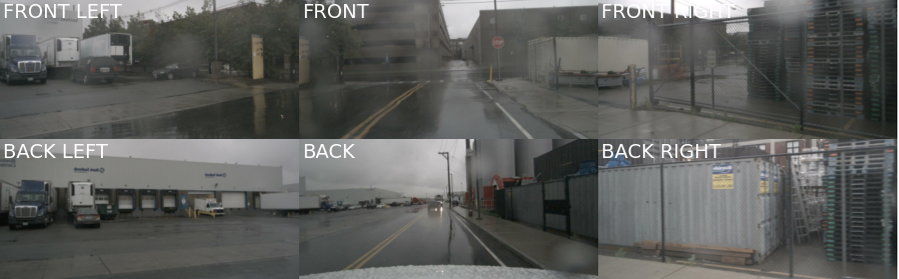}\\
        \includegraphics[width=0.993\textwidth,height=0.9in]{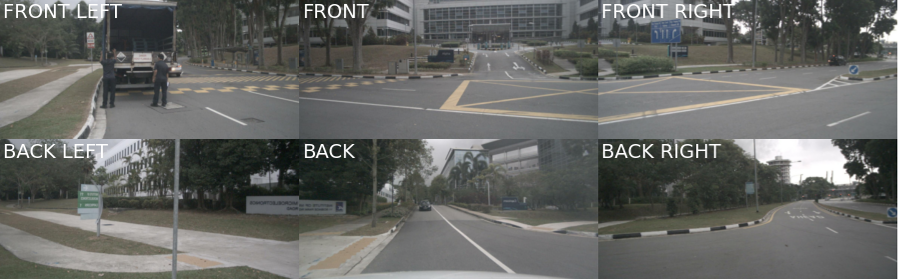}\\
        \includegraphics[width=0.993\textwidth,height=0.9in]{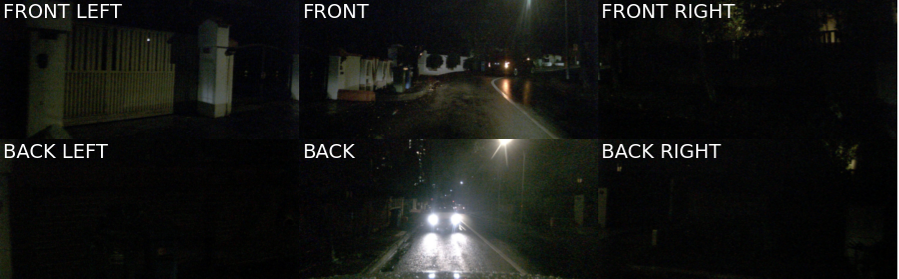}\\
\end{minipage}%
}%
\subfigure[Static]{
\begin{minipage}{0.149\linewidth}
        \centering
        \includegraphics[width=0.993\textwidth,height=0.9in]{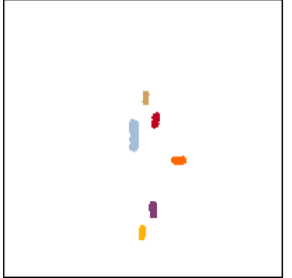}\\
        \includegraphics[width=0.993\textwidth,height=0.9in]{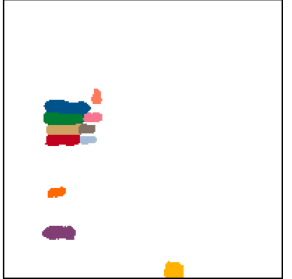}\\
        \includegraphics[width=0.993\textwidth,height=0.9in]{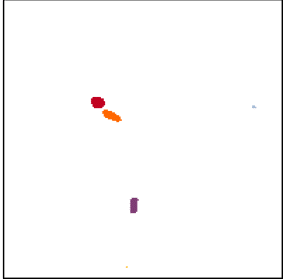}\\
        \includegraphics[width=0.993\textwidth,height=0.9in]{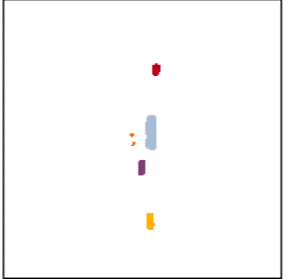}\\
\end{minipage}%
}%
\subfigure[Fiery~\cite{hu2021fiery}]{
\begin{minipage}{0.149\linewidth}
        \centering
        \includegraphics[width=0.993\textwidth,height=0.9in]{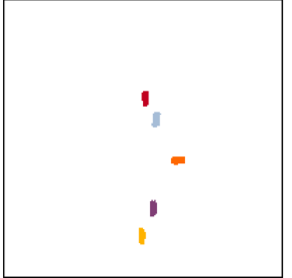}\\
        \includegraphics[width=0.993\textwidth,height=0.9in]{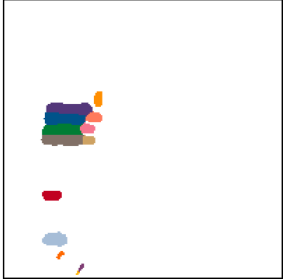}\\
        \includegraphics[width=0.993\textwidth,height=0.9in]{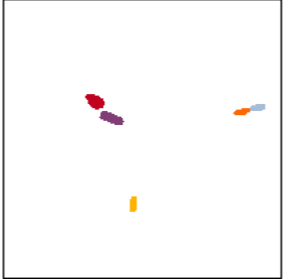}\\
        \includegraphics[width=0.993\textwidth,height=0.9in]{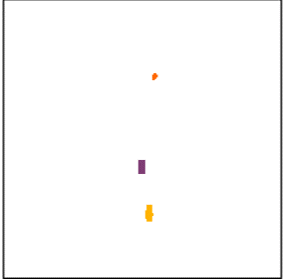}\\
\end{minipage}%
}%
\subfigure[ours]{
\begin{minipage}{0.149\linewidth}
        \centering
        \includegraphics[width=0.993\textwidth,height=0.9in]{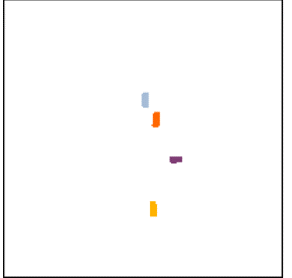}\\
        \includegraphics[width=0.993\textwidth,height=0.9in]{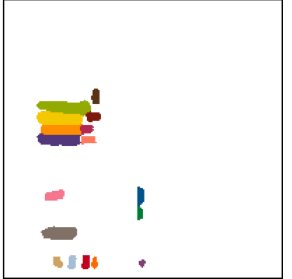}\\
        \includegraphics[width=0.993\textwidth,height=0.9in]{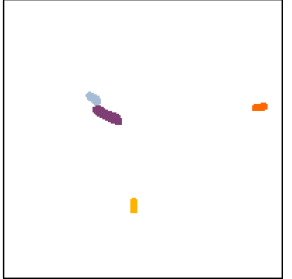}\\
        \includegraphics[width=0.993\textwidth,height=0.9in]{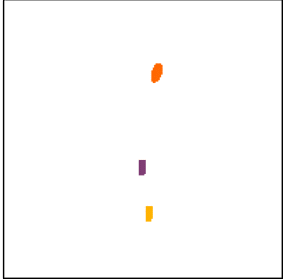}\\
\end{minipage}%
}%
\subfigure[GT]{
\begin{minipage}{0.149\linewidth}
        \centering
        \includegraphics[width=0.993\textwidth,height=0.9in]{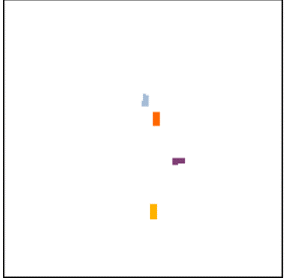}\\
        \includegraphics[width=0.993\textwidth,height=0.9in]{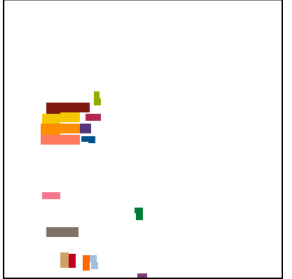}\\
        \includegraphics[width=0.993\textwidth,height=0.9in]{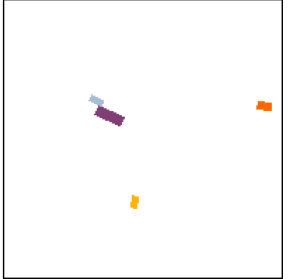}\\
        \includegraphics[width=0.993\textwidth,height=0.9in]{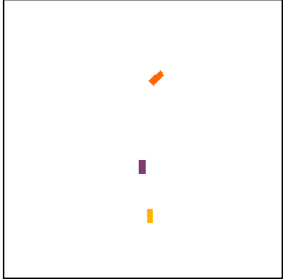}\\
\end{minipage}%
}%
\caption{Qualitative results on BEV semantic segmentation. The resolution setting is 100m × 100m at 50cm resolution. For the best view,
the cars is marked in instance level.}
\label{fig:car}
\end{figure*}

\section{Experiments}
\subsection{Dataset}
We evaluate our method on the NuScenes dataset~\cite{caesar2020nuscenes}, which includes a fully autonomous vehicle sensor suite with six cameras, five radars, and one lidar, providing a complete 360-degree field of view. The dataset contains 1,000 diverse scenes in various weather conditions, times of day, and traffic situations, divided into 700/150/150 for training/validation/testing. Each scene lasts 20 seconds and is annotated at 2Hz, totaling 40,000 samples. The camera's intrinsic and extrinsic matrices are available for every scene.

Following previous works~\cite{hu2021fiery, hu2022stp3}, we focus on the BEV segmentation task, i.e., semantic segmentation, instance segmentation, and perceived maps prediction. We mainly evaluate our model using the IoU metric and also employ panoptic quality (PQ), recognition quality (RQ), and segmentation quality (SQ) to measure instance segmentation quality, following metrics in the video prediction area~\cite{kim2020video}.

\subsection{Implementation Details}
Our framework is implemented in PyTorch~\cite{paszke2019pytorch}. We set the temporal field to 5, corresponding to 2.0s of past content in the NuScenes dataset. For future instance segmentation, our model predicts 2.0s in the future, which corresponds to 4 frames. The camera images are resized to $224 \times 480$, while the output spatial dimension of our model is fixed at $200 \times 200$. The default resolution in both the x and y directions of prediction is $100m \times 100m$ at 50cm pixel resolution, unless specified otherwise.

Like~\cite{hu2021fiery, hu2022stp3}, we use a pre-trained EfficientNet-B4~\cite{tan2019efficientnet} as the backbone. The Adam optimizer with a constant learning rate of $1 \times 10^{-3}$ is used for semantic segmentation and perceived maps prediction, while $3 \times 10^{-4}$ is used for future instance segmentation. We train our model on 8 Tesla V100 GPUs with a batch size of 16 for 20 epochs using mixed precision.

\subsection{Experimental Results}


\begin{table}[t]
\centering
\caption{Bird’s-eye view semantic segmentation of the present timeframe on NuScenes dataset.
Setting 1: 100m × 50m at 25cm resolution. Setting 2: 100m × 100m at 50cm resolution. Setting 3: 32.0m × 19.2m at 10cm resolution. }
\label{tab:sem seg}
\resizebox{\linewidth}{!}{%
\begin{tabular}{lccc}
\toprule
\multirow{2}{*}{Method} & \multicolumn{3}{c}{Intersection-over-Union (IoU)} \\\cmidrule{2-4}
 & Setting 1 & Setting 2 & Setting 3 \\
\midrule
VED~\cite{lu2019monocular} & 8.8 & - & - \\
PON~\cite{roddick2020predicting} & 24.7 & - & - \\
VPN~\cite{pan2020cross} & 25.5 & - & - \\
STA~\cite{saha2021enabling} & 36.0 & - & - \\
Lift-Splat~\cite{philion2020lift} & - & 32.1 & - \\
Fishing Camera~\cite{hendy2020fishing} & - & - & 30.0 \\
zhou~\etal~\cite{zhou2022cross} & 37.5 & 36.0 & - \\
FIERY~\cite{hu2021fiery} & 39.9 & 38.2 & 57.6 \\
Geo-ConvGRU(ours) & \textbf{41.7} & \textbf{39.5} & \textbf{59.3}\\
\bottomrule
\end{tabular}%
}
\vspace{-0.4cm}
\end{table}

\begin{figure*}[t]
\centering
\subfigure[Image]{
\begin{minipage}{0.39\linewidth}
        \centering
        \includegraphics[width=0.993\textwidth,height=0.9in]{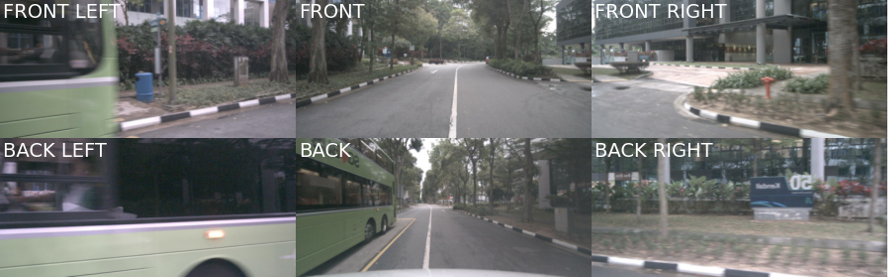}\\
        \includegraphics[width=0.993\textwidth,height=0.9in]{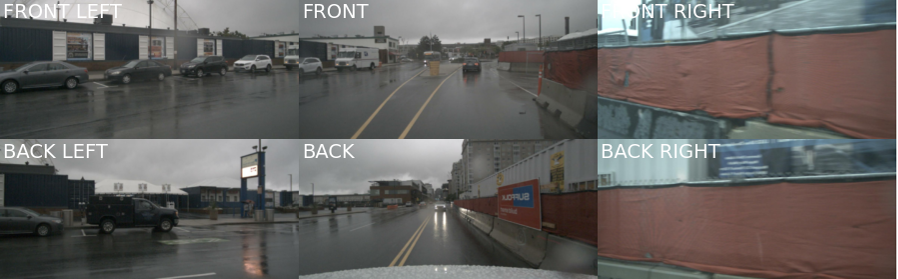}\\
        \includegraphics[width=0.993\textwidth,height=0.9in]{images/car/car_img_4}\\

\end{minipage}%
}%
\subfigure[ST-P3~\cite{hu2022stp3}]{
\begin{minipage}{0.149\linewidth}
        \centering
        \includegraphics[width=0.993\textwidth,height=0.9in]{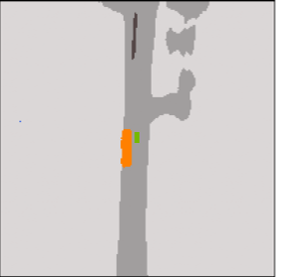}\\
        \includegraphics[width=0.993\textwidth,height=0.9in]{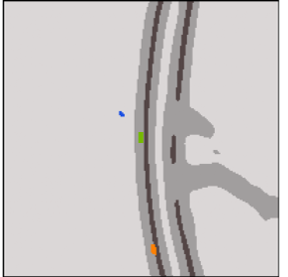}\\
        \includegraphics[width=0.993\textwidth,height=0.9in]{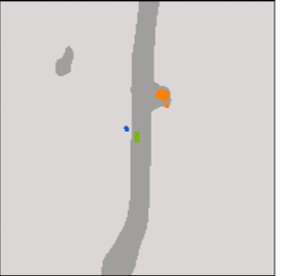}\\
\end{minipage}%
}%
\subfigure[ours]{
\begin{minipage}{0.149\linewidth}
        \centering
        \includegraphics[width=0.993\textwidth,height=0.9in]{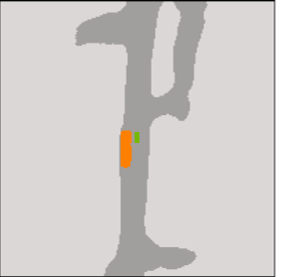}\\
        \includegraphics[width=0.993\textwidth,height=0.9in]{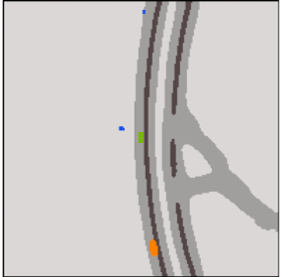}\\
        \includegraphics[width=0.993\textwidth,height=0.9in]{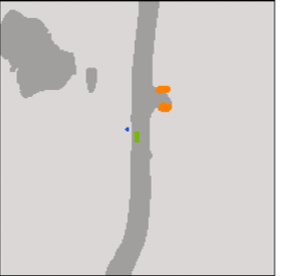}\\

\end{minipage}%
}%
\subfigure[GT]{
\begin{minipage}{0.149\linewidth}
        \centering
        \includegraphics[width=0.993\textwidth,height=0.9in]{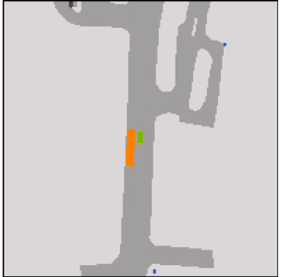}\\
        \includegraphics[width=0.993\textwidth,height=0.9in]{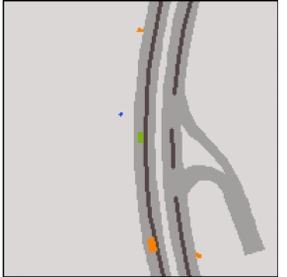}\\
        \includegraphics[width=0.993\textwidth,height=0.9in]{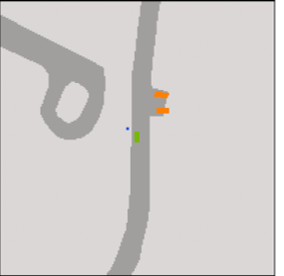}\\

\end{minipage}%
}%
\caption{Qualitative results on perceived maps prediction .}
\label{fig:per}
\end{figure*}

\begin{table*}[]
\centering
\caption{ Bird’s-eye view perceived maps for on NuScenes dataset.$*$ means results come from re-implementation.}
\label{tab:perceived maps}
\resizebox{0.6\textwidth}{!}{%
\begin{tabular}{lccccc}
\toprule
Method & Drivable Area & Lane & Vehicle & Pedestrian & avg \\
\midrule
VED~\cite{lu2019monocular} & 60.8 & 16.7 & 23.3 & 11.9 & 28.2 \\
VPN~\cite{pan2020cross} & 66.0 & 17.1 & 28.2 & 10.3 & 30.4 \\
PON~\cite{roddick2020predicting} & 63.1 & 17.2 & 27.9 & 13.9 & 30.5 \\
Lift-Splat~\cite{philion2020lift} & 72.2 & 20.0 & 31.2 & 15.0 & 34.6 \\
IVMP~\cite{wang2021learning} & 74.7 & 20.9 & 34.0 & 17.4 & 36.8 \\
FIERY~\cite{hu2021fiery} & 72.0 & 33.6 & 38.0 & 17.2 & 40.2 \\
ST-P3*~\cite{hu2022stp3} & 72.7 & 37.1 & 38.6 & 16.9 & 41.3 \\
Geo-ConvGRU(ours) & \textbf{73.5} & \textbf{37.6} & \textbf{39.6} & \textbf{17.5} & \textbf{42.1}\\
\bottomrule
\end{tabular}%
}
\end{table*}

\par\noindent\textbf{Result on BEV semantic segmentation.} We compare Geo-ConvGRU with state-of-the-art methods, including VED~\cite{lu2019monocular}, PON~\cite{roddick2020predicting}, VPN~\cite{pan2020cross}, STA~\cite{saha2021enabling}, Lift-Splat~\cite{philion2020lift}, Fishing Camera~\cite{hendy2020fishing}, zhou~\etal~\cite{zhou2022cross}, and FIERY~\cite{hu2021fiery}. Following FIERY~\cite{hu2021fiery}, three different resolution settings are chosen. In detail, Setting 1 adopts 100m$\times$ 50m area around the vehicle and samples a map at a 25cm resolution. Setting 2 adopts 100m$\times$ 100m area around the vehicle and samples a map at a 50cm resolution. Setting 3 adopts 32m$\times$19.2m area around the vehicle and samples a map at a 10cm resolution. We report the mIoU performance for all methods in Table~\ref{tab:sem seg}. Note that, for a fair comparison, the EfficientNet-B4~\cite{tan2019efficientnet} is chosen as the shared backbone. Unsurprisingly, our Geo-ConvGRU achieves \textbf{1.8\%}, \textbf{1.3\%}, and \textbf{1.6\%} gains in terms of mIoU over the best score obtained by the strongest existing method (FIERY). At the same time, we also display the qualitative results on 100m × 100m at 50cm resolution in Fig~\ref{fig:car}. Compared with static results, the methods where a temporal module is added get better results and visualization performance. Moreover, compared with FIERY~\cite{hu2021fiery}, owing to adding a geographical mask, our Geo-ConvGRU can successfully avoid wrong predictions on some wrong mobile cars. For example, according to the first and third rows in Fig~\ref{fig:car}, the moving cars in the previous frame are still marked in the current frame.

\par\noindent\textbf{Result on perceived maps prediction.} Similar to BEV semantic segmentation, we compare our method with state-of-the-art methods for the perceived maps prediction task. Except with the mentioned methods in the BEV semantic segmentation task, we also compare with ST-P3~\cite{hu2022stp3} and IVMP~\cite{wang2021learning}, which are unique designs for this task. Note that the result of ST-P3 in table~\ref{tab:perceived maps} is based on our re-implementation from the ST-P3's official code. The resolution setting is the setting 2 in BEV semantic segmentation. Compared with our baseline FIERY~\cite{hu2021fiery}, our method achieves \textbf{1.9\%} performance improvement on average for all classes. Moreover, our method's performance in each class outperforms the new state-of-the-art method, ST-P3. Furthermore, the qualitative results on perceived maps prediction are shown in Fig.~\ref{fig:per}. Compared with ST-P3, our method generates a more accurate perceived map, especially for the drivable area class.

\begin{table}[]
\centering
\caption{Bird’s-eye view future instance segmentation for 2.0s on NuScenes dataset. We report future semantic segmentation IoU (\%) and instance segmentation metrics from video prediction area.
PQ, SQ and RQ are short for panoptic quality, recognition qualityand segmentation quality, respectively. {The static method assumes all obstacles static in the prediction horizon.} 
$*$ means results come from re-implementation.}
\label{tab:ins seg}
\resizebox{\linewidth}{!}{%
\begin{tabular}{lcccc}
\toprule
\multirow{2}{*}{Method} & Future Semantic Seg. & \multicolumn{3}{c}{Future Instance Seg.} \\ \cmidrule{2-5}
 & IoU & PQ & SQ & RQ \\
\midrule
Static & 32.2 & 27.6 & 70.1 & 39.1 \\
FIERY*~\cite{hu2021fiery} & 36.1 & 27.4 & 69.4 & 39.5 \\
ST-P3*~\cite{hu2022stp3} & 36.8 & 28.4 & 69.3 & 41.0 \\
Geo-ConvGRU(ours) & \textbf{37.7} & \textbf{29.8} & \textbf{70.3} & \textbf{42.7}\\
\bottomrule
\end{tabular}%
}
\vspace{-0.4cm}
\end{table}

\begin{figure*}[]
\centering
\subfigure[Image]{
\begin{minipage}{0.59\linewidth}
        \centering
       \includegraphics[width=0.993\textwidth,height=0.9in]{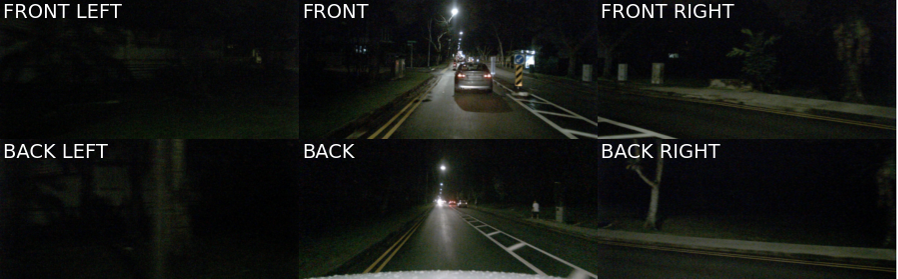}\\
        \includegraphics[width=0.993\textwidth,height=0.9in]{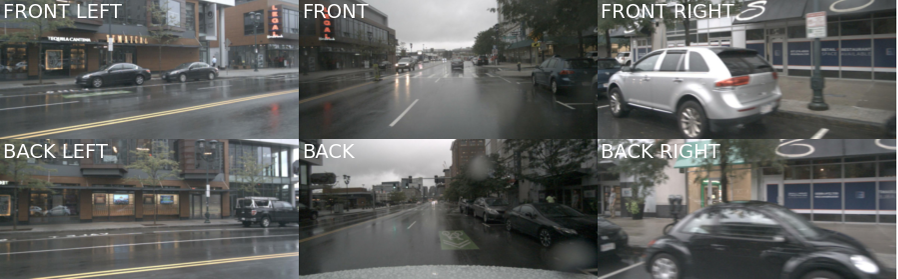}\\
       \includegraphics[width=0.993\textwidth,height=0.9in]{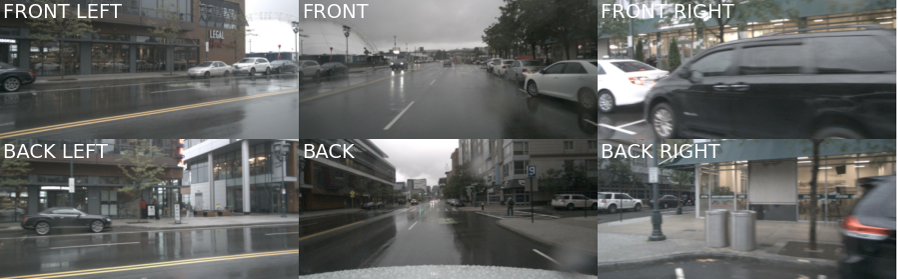}\\
       \includegraphics[width=0.993\textwidth,height=0.9in]{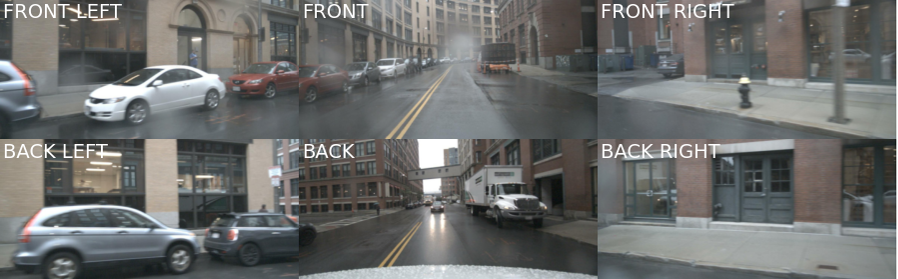}\\
       \includegraphics[width=0.993\textwidth,height=0.9in]{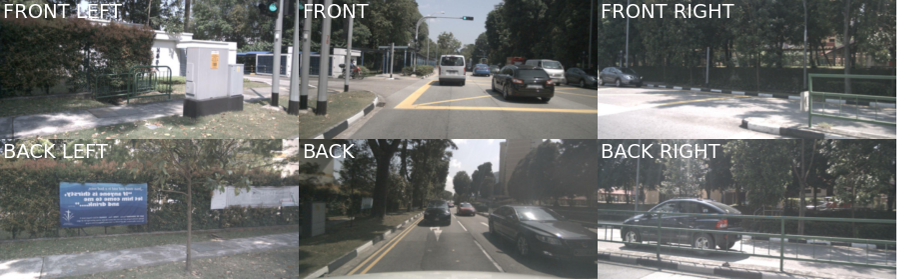}\\

\end{minipage}%
}%
\subfigure[ours]{
\begin{minipage}{0.19\linewidth}
        \centering
       \includegraphics[width=0.993\textwidth,height=0.9in]{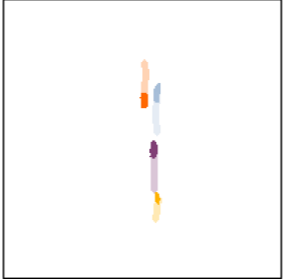}\\
        \includegraphics[width=0.993\textwidth,height=0.9in]{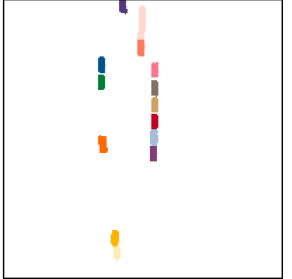}\\
       \includegraphics[width=0.993\textwidth,height=0.9in]{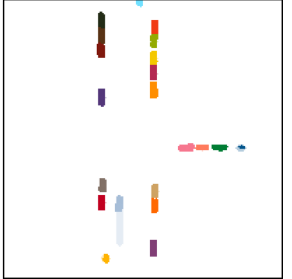}\\
       \includegraphics[width=0.993\textwidth,height=0.9in]{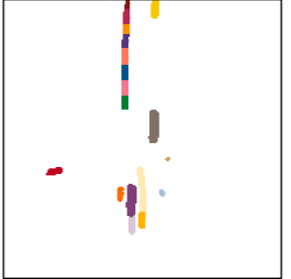}\\
       \includegraphics[width=0.993\textwidth,height=0.9in]{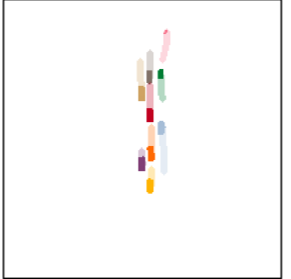}\\

\end{minipage}%
}%

\caption{Qualitative results on future instance segmentation. The furture frame predictions are marked by lighter color for best view.}
\label{fig:pred}
\end{figure*}

\par\noindent\textbf{Result on future instance segmentation.} To further evaluate our method, we compare the performance of Geo-ConvGRU for future instance segmentation. The model trained for future instance segmentation relies on the static model. For a fair comparison, we train our model, FIERY, and ST-P3 with the same static model. Therefore, we use the results from re-implementation in Table~\ref{tab:ins seg}. Like BEV semantic segmentation and perceived map prediction, Geo-ConvGRU achieves the new state-of-the-art future instance segmentation. It exceeds the ST-P3 by \textbf{0.9\%} for the semantic segmentation matrix, IoU. Moreover, Geo-ConvGRU also gets 1.4\%, 1.0\%, and 1.7\% for instance segmentation metrics, i.e., PQ, SQ, RQ. The qualitative results on future instance segmentation are shown in Fig~\ref{fig:pred}.
Geo-ConvGRU has demonstrated its ability to make accurate future instance segmentation predictions regardless of unfavorable weather conditions. This suggests that the network's robustness is enhanced by the use of Geo-ConvGRU.

\begin{table*}[]
\centering
\caption{Ablation study on Geo-ConvGRU.T is short for temporal filed.  Short and long denote 30m × 30m and 100m × 100m around the ego-vehicle, respectively.}
\label{tab:ab}
\begin{tabular}{lccccccc}
\toprule
\multirow{2}{*}{Method} & \multirow{2}{*}{{T}} & \multicolumn{1}{l}{\multirow{2}{*}{{Train Time(h)}}} & \multicolumn{1}{l}{\multirow{2}{*}{{FPS(Hz)}}} & \multicolumn{2}{c}{IoU} & \multicolumn{2}{c}{PQ} \\ \cline{5-8}
 &  & \multicolumn{1}{l}{} & \multicolumn{1}{l}{} & short & long & short & long \\
\midrule
baseline & 3 & 21.6 & 7.9 & 67.2 & 37.7 & 58.6 & 30.6 \\
TADA\cite{huang2021tada} & 3 & 17.8 & 7.5 & 38.8 & 14.7 & 25.5 & 9.7 \\
SimVP\cite{gao2022simvp} & 3 & 21.1 & 7.3 & 65.4 & 36.9 & 55.9 & 28.8 \\
ConvLSTMcite\cite{shi2015convolutional} & 3 & 20.9 & 6.2 & 67.1 & 35.8 & 57.9 & 30.4 \\
PredRNN\cite{wang2022predrnn} &3 & 21.2 & 5.9 & 49.5 & 28.3 & 38.3 & 19.0 \\
PredRNN++\cite{wang2018predrnn++} &3 & 20.5 & 6.2 & 49.6 & 27.7 & 37.8 & 18.8 \\
BEVformer & 3 & 64.3 & 1.9 & 67.4 & 39.0 & 59.5 & \textbf{32.7} \\
\midrule
ConvGRU & 3 & 19.5 & 6.5 & 67.4 & 38.2 & 58.5 & 31.5 \\
\multirow{3}{*}{Geo-ConvGRU} & 3 & 19.8 & 6.4 & 66.8 & 38.8 & 57.2 & 32.1 \\
 & 5 & 26.2 & 4.9 & 68.6 & 39.5 & \textbf{59.9} & 31.9\\
  & 7 & 33.9 & 3.3 & \textbf{69.2} & \textbf{39.8} & \textbf{59.9} & 32.2\\
\bottomrule
\end{tabular}%
\end{table*}

\begin{figure*}[t]
\centering
\subfigure[Image]{
\begin{minipage}{0.39\linewidth}
        \centering
        \includegraphics[width=0.993\textwidth,height=0.9in]{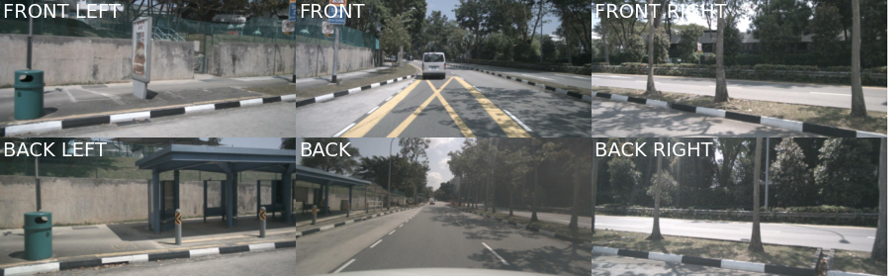}\\
        \includegraphics[width=0.993\textwidth,height=0.9in]{images/car/car_img_5}\\
\end{minipage}%
}%
\subfigure[ConvGRU]{
\begin{minipage}{0.149\linewidth}
        \centering
        \includegraphics[width=0.993\textwidth,height=0.9in]{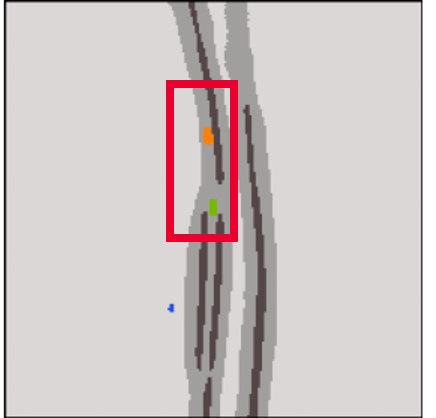}\\
        \includegraphics[width=0.993\textwidth,height=0.9in]{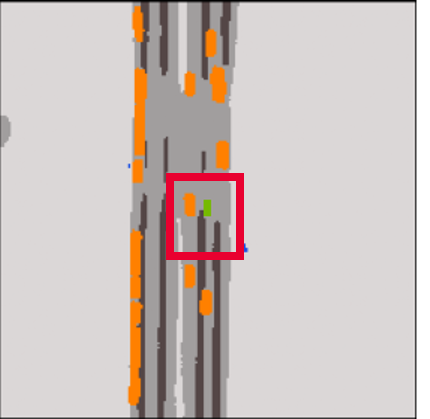}\\
\end{minipage}%
}%
\subfigure[ours]{
\begin{minipage}{0.149\linewidth}
        \centering
        \includegraphics[width=0.993\textwidth,height=0.9in]{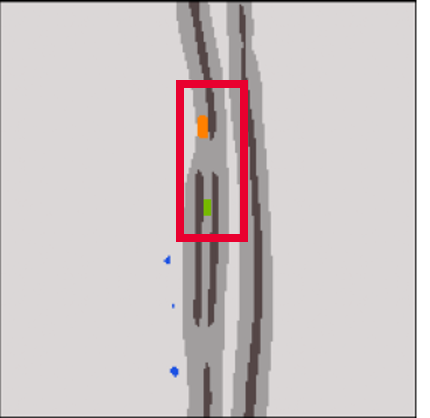}\\
        \includegraphics[width=0.993\textwidth,height=0.9in]{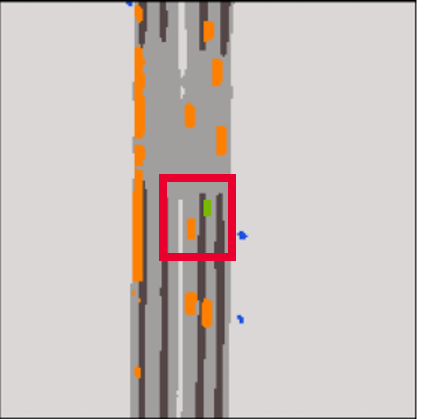}\\
\end{minipage}%
}%
\subfigure[GT]{
\begin{minipage}{0.149\linewidth}
        \centering
        \includegraphics[width=0.993\textwidth,height=0.9in]{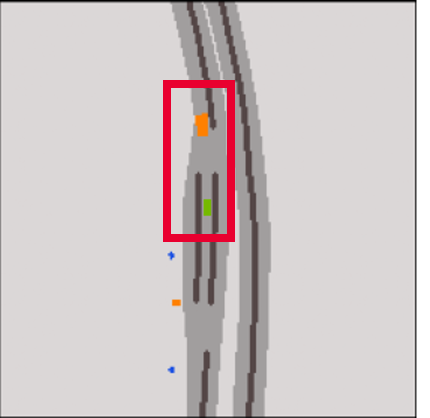}\\
        \includegraphics[width=0.993\textwidth,height=0.9in]{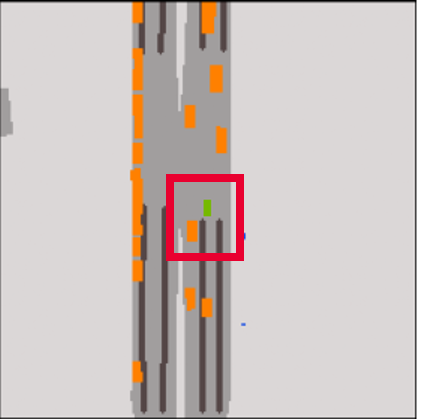}\\
\end{minipage}%
}%
\caption{Visulization results of ConvGRU and Geo-ConvGRU on perceived maps prediction.}
\label{fig:ab per}
\end{figure*}


\subsection{Ablation study}

First, we employ various state-of-the-art temporal/spatial-temporal methods as a temporal module to compare with our Geo-ConvGRU in Table~\ref{tab:ab}. These methods include TADA~\cite{huang2021tada}, SimVP~\cite{gao2022simvp}, ConvLSTM~\cite{shi2015convolutional}, PredRNN~\cite{wang2022predrnn},  PredRNN++\cite{wang2018predrnn++}, and BEVFormer~\cite{li2022bevformer}. 
{ The 3D convolution from Fiery~\cite{hu2021fiery} is utilized as the baseline for this study. To ensure a fair comparison, the temporal field remains fixed at 3. According to the top section of Table~\ref{tab:ab}, ConvGRU significantly outperforms other methods, except for BEVFormer~\cite{li2022bevformer}. However, integrating a transformer structure results in considerable increases in training time and a marked decrease in FPS, which are deemed unacceptable. When considering the trade-off between accuracy and efficiency, ConvGRU emerges as a more viable selection.} It can solidly confirm our argument that ConvGRU is an effective method to get better performance for BEV tasks as the video representation task. We also compare ConvGRU and Geo-ConvGRU in Table~\ref{tab:ab}. There is a reasonable improvement by adding a geographical mask into ConvGRU. 
{
Furthermore, the outcomes of long time-series tasks are presented at the bottom of Table~\ref{tab:ab}. Notably, the performance of Geo-ConvGRU improves with the addition of more previous frames. Nevertheless, long-time series result in increased training time and slower FPS, unsuitable for the real-time demands of autonomous driving. Balancing accuracy and efficiency demands, a temporal field setting of 5 is deemed optimal.
}
Meanwhile, we show ConvGRU and Geo-ConvGRU visualization results on perceived maps prediction in Fig.~\ref{fig:ab per}. Similar to the situation we explain in BEV semantic segmentation, ConvGRU is prone to making wrong predictions on some mobile cars owing to adding the temporal module. A geographical mask can significantly solve this issue by weighting the temporal module's output. According to Fig.\ref{fig:ab per} red box, Geo-ConvGRU provides a more accurate prediction of nearby moving cars than ConvGRU. It supports that adding a geographical mask can relieve the problem because fusing spatial and temporal features model is prone to overfitting at moving pixels.

\section{Conclusions}

In this paper, we presented Geo-ConvGRU, a novel approach to address the BEV semantic segmentation, perceived maps prediction, and future instance segmentation tasks. Our method effectively combines spatial and temporal features using a ConvGRU-based architecture while incorporating a geographical mask to alleviate issues arising from moving objects in the scene. We have demonstrated that Geo-ConvGRU consistently outperforms state-of-the-art methods on all three tasks, achieving substantial improvements in terms of mIoU, PQ, SQ, and RQ metrics. Furthermore, we conducted a thorough ablation study to validate the effectiveness of the ConvGRU and the geographical mask components in our proposed approach. The results confirm that the ConvGRU is a suitable choice for handling the temporal aspect of these tasks, and the geographical mask helps in improving the model's performance by mitigating overfitting issues related to moving objects.

In conclusion, our work contributes a novel and effective approach for handling BEV semantic segmentation, perceived maps prediction, and future instance segmentation tasks. We believe that our findings and insights will inspire future research in this area, paving the way for more robust and accurate autonomous driving systems.

\ifCLASSOPTIONcompsoc
\section*{Acknowledgments}
\else
\section*{Acknowledgment}
\fi

This work was done during Guanglei Yang's internship at Huawei. 
\bibliographystyle{IEEEtran}
\bibliography{egbib}

\end{document}